\documentclass[letterpaper,10pt,conference]{ieeeconf}
\IEEEoverridecommandlockouts

\usepackage{mathptmx}   
\usepackage{amsfonts, amssymb, yhmath}
\usepackage{epsfig}
\usepackage{theorem}
\usepackage{algorithmic}
\usepackage{algorithm}
\usepackage{textcomp}
\usepackage{multirow}

\newtheorem{theorem}{Theorem}

\newtheorem{proposition}[theorem]{Proposition}
\newtheorem{corollary}[theorem]{Corollary}

\newtheorem{pro}[theorem]{Problem}

\newcommand{\qed}{\hfill $\Box$\\}

\def\tompp{{\sc Tompp}}
\def\dompp{{\sc Dompp}}

\def\mpp{\textrm{MPP}}

\begin{document}

\title{Planning Optimal Paths for Multiple Robots on Graphs}
\author{
\begin{tabular}{ccc}
Jingjin Yu & & Steven M. LaValle 
\end{tabular}
\thanks{Jingjin Yu is with the Department of Electrical and Computer Engineering, University of Illinois at Urbana-Champaign, Urbana, IL 61801 USA. E-mail: jyu18@uiuc.edu. Steven M. LaValle is with the Department of Computer Science, University of Illinois at Urbana-Champaign, Urbana, IL 61801 USA. E-mail: lavalle@uiuc.edu. This work was supported in part by NSF grants 0904501 (IIS Robotics) and 1035345 (Cyberphysical Systems), DARPA SToMP grant HR0011-05-1-0008, and MURI/ONR grant N00014-09-1-1052.}
}
\maketitle

\begin{abstract} In this paper, we study the problem of optimal multi-robot path planning ($\mpp$) on graphs. We propose two multiflow based integer linear programming (ILP) models that computes minimum last arrival time and minimum total distance solutions for our $\mpp$ formulation, respectively. The resulting algorithms from these ILP models are complete and guaranteed to yield true optimal solutions. In addition, our flexible framework can easily accommodate other variants of the $\mpp$ problem. Focusing on the time optimal algorithm, we evaluate its performance, both as a stand alone algorithm and as a generic heuristic for quickly solving large problem instances. Computational results confirm the effectiveness of our method.\end{abstract}

\section{Introduction}
\label{sec:intro}
Planning collision-free paths for multiple robots, an easily stated yet difficult problem, has been actively studied for decades \cite{ErdLoz86, LavHut98b, LunBer11, Rya08, Sil05, StaKor11, Sur09, BerOve05, BerSnoLinMan09, Zel92}. The hardness of the problem mainly resides with the coupling between the robots' paths which leads to an enormous state space and branching factor. As such, algorithms that are both complete and (distance) optimal, such as the A$^*$ \cite{HarNilRap68} algorithm and its variants, do not perform well on tightly coupled problems beyond very small ones. On the other hand, faster algorithms for finding the paths generally do not provide optimality guarantees: Sifting through all feasible path sets for optimal ones greatly increases the search space, which often makes these problems intractable. 

In this paper, we investigate the problem of planning optimal paths for multiple robots with individual goals. The robots have identical but non-negligible sizes, are confined to some arbitrary connected graph, and are capable of moving from one vertex to an adjacent vertex in one time step. Collision between robots is not allowed, which may occur when two robots attempt to move of the same vertex or move along the same edge in different directions. For this general setting, we propose a network flow based integer linear programming (ILP) model for finding robot paths that are time optimal or distance optimal. Our time optimality criterion seeks to minimize the number of time steps until the last robot reaches its goal; distance optimality seeks to minimize the total distance (each edge has unit distance) traveled by the robots. Taking advantage of the state of the art ILP solvers (Gurobi is used in this paper), our method can plan time optimal, collision-free paths for several dozens of robots on graphs with hundreds of vertices within minutes. 

As a universal subroutine, collision-free path planning for multiple robots finds applications in tasks spanning assembly \cite{HalLatWil00, Nna92}, evacuation \cite{RodAma10}, formation control \cite{BalArk98, PodSuk04, ShuMurBen07,  SmiEgeHow08, TanPapKum04}, localization \cite{FoxBurKruThr00}, object transportation \cite{MatNilSim95, RusDonJen95}, search and rescue \cite{JenWheEva97}, and so on. Given its importance, path planning for multi-robot systems has remained as a subject of intense study for many decades. Given the vast size of the available literature, we will only mention related research on discrete $\mpp$ and refer the readers to \cite{ChoLynHutKanBurKavThr05, Lat91, Lav06} and the references therein for a more comprehensive review of the subject. 

From an algorithmic perspective, discrete $\mpp$ is a natural extension of the single robot path planning problem: One may combine the state spaces of all robots and treat the problem as a planning problem for a single robot. A$^*$ algorithm can then be used to compute distance optimal solutions to these problems. However, since naive A$^*$ scales poorly due to the curse of dimensionality, additional heuristic methods were proposed to improve the computational performance. One of the first such heuristics, Local Repair A$^*$ (LRA$^*$) \cite{Zel92}, plans robot paths simultaneously and performs local repairs when conflicts arise. Focusing on fixing the (locality) shortcomings of LRA$^*$, Windowed Hierarchical Cooperative A$^*$ (WHCA$^*$) \cite{Sil05} proposed to use a space-time window to allow more choices for resolving local conflicts while limiting the search space size at the same time. For additional heuristics exploring various specific local and global features, see \cite{LunBer11, Rya08, Sur09}. 

Formulations of $\mpp$ problems with optimality guarantee have also been studied. The most general optimality criterion is the total path length traveled by all robots, which is consistent with the distance heuristic used by the A$^*$ algorithm. Since A$^*$ is the best possible among all such algorithms for finding distance optimal solutions, one should not expect complete and true optimal algorithms to exist that perform much better than the basic A$^*$ algorithm in all cases. Nevertheless, this does not prevent algorithms from quickly solving certain instances optimally. One such algorithm that is also complete, MGS$x$, is presented in \cite{StaKor11} (note that the grid world formulation in \cite{StaKor11}, which allows diagonal moves in general, even in the presence of diagonal obstacles, does not carry over to general graphs or geometric models in robotics). For time optimality, for a version of the $\mpp$ problem that resembles our formulation more closely, it was shown that finding a time optimal solution is NP-hard \cite{Sur10}, implying that our formulation is also intractable \cite{YuArxiv-1205-5263}. Finally, it was shown that finding the least number of moves for the $N\times N$-generalization of the 15-puzzle is NP-hard \cite{RatWar90}. Here, time optimality equals distance optimality, which is not the case in general. 

The main contributions of this paper are twofold. First, adapting the constructions from \cite{YuLav12WAFR-A}, we develop ILP models for solving time optimal and distance optimal $\mpp$ problems. The resulting algorithms are shown to be complete. Our approach is quite general and easily accommodates other formulations of the $\mpp$ problems, including that of \cite{StaKor11}. Second, we provide thorough computational evaluations of our models' performance: With a state-of-the-art ILP solver, our models are capable of solving large problem instances with few dozens of robots fairly fast. Such a result is in some sense the best we can hope for because the best possible algorithm for such problems cannot run in polynomial time unless $P = NP$. As an added bonus, we also show that the (time optimal) algorithm works well as a subroutine for quickly solving $\mpp$ problems (non-optimally)\footnote{The software (written in Java, including a programming interface), as well as all examples used in our evaluation, are available at \texttt{http://msl.cs.uiuc.edu/{\texttildelow}jyu18/pe/mapp.html}.}.

The rest of the paper is organized as follows. We provide problem definitions in Section \ref{sec:definition}, along with a motivating example. Section \ref{sec:planning-and-flow} relates $\mpp$ to multiflow, establishing the equivalence between the two problems. In Section \ref{sec:algorithm}, ILP models are provided for obtaining time optimal and distance optimal solutions, respectively. Section \ref{sec:puzzle} is devoted to briefly discussing basic properties of the $n^2$-puzzle, which is an interesting benchmark problem on its own. We evaluate the computational performance of our algorithm in Section \ref{sec:evaluation} and conclude in Section \ref{sec:conclusion}.

\section{Multi-robot Path Planning on Graphs}\label{sec:definition}
\subsection{Problem Formulation}
Let $G = (V, E)$ be a connected, undirected, simple graph (i.e., no multi-edges), in which $V = \{v_i\}$ is its vertex set and $E = \{(v_i, v_j)\}$ is its edge set. Let $R = \{r_1, \ldots, r_n\}$ be a set of robots that move with unit speeds along the edges of $G$, with initial and goal locations on $G$ given by the injective maps $x_I, x_G: R \to V$, respectively. The set $R$ is effectively an index set. A {\em path} or {\em scheduled path} is a map $p_i: \mathbb Z^+ \to V$, in which $\mathbb Z^+ := \mathbb N \cup \{0\}$. Intuitively, the domains of the paths are discrete time steps. A path $p_i$ is {\em feasible} for a single robot $r_i$ if it satisfies the following properties: 1. $p_i(0) = x_I(r_i)$; 2. For each $i$, there exists a smallest $k_i^{\min} \in \mathbb Z^+$ such that for all $k \ge k_i^{\min}$, $p_i(k) \equiv x_G(r_i)$; 3. For any $0 \le k < k_i^{\min}$, $(p_i(k), p_i(k+1)) \in E$ or $p_i(k) = p_i(k+1)$. We say that two paths $p_i, p_{j}$ are in {\em collision} if there exists $k \in \mathbb Z^+$ such that $p_i(k) = p_{j}(k)$ (collision on a vertex, or {\em meet}) or $(p_i(k), p_i(k+1)) = (p_j(k+1), p_j(k))$ (collision on an edge, or {\em head-on}). If $p(k) = p(k+1)$, then the robot stays at vertex $p(k)$ between the time steps $k$ and $k+1$. 

\begin{pro}[$\mpp$ on Graphs]\label{pimpp} Given $(G, R, x_I, x_G)$, find a set of paths $P = \{p_1, \ldots, p_n\}$ such that $p_i$'s are feasible paths for respective robots $r_i$'s and no two paths $p_i, p_j$ are in collision. 
\end{pro}

\begin{figure}[htp]
\begin{center}
    \includegraphics[width=0.16\textwidth]{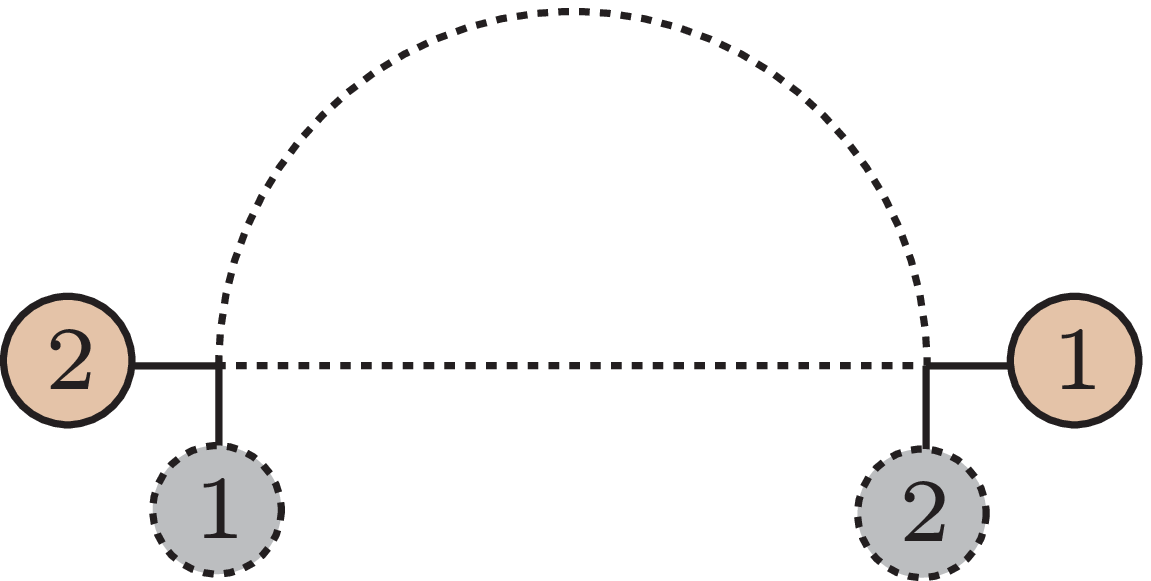} 
\end{center}
\vspace*{-1mm}
\caption{\label{fig:optimality} Time optimality and distance optimality cannot be satisfied simultaneously for this setup.}
\end{figure}

A natural criterion for measuring path set optimality is the number of time steps until the last robot reaches its goal. This is sometimes called the {\em makespan}, which can be computed from $\{k_i^{\min}\}$ for a feasible path set $P$ as 
\begin{displaymath}
T_P = \max_{1 \le i \le n}k_i^{\min}.
\end{displaymath}
Another frequently used objective is distance optimality, which counts the total number of edges traveled by the robots. We point out that distance optimality and time optimality cannot be satisfied at the same time in general: In Fig. \ref{fig:optimality}, let the dotted straight line have length $t$ and the dotted arc has length $1.5t$ from some large even number $t$. The four solid line segments are edges with unit length. Assuming that robot 1, 2 are to move from the locations marked with solid circles to the locations marked with gray dotted circles. Time optimal paths take $1.5t + 2$ time steps with a total distance of $2.5t + 4$; distance optimal paths take $2t + 3$ time steps with a total distance of $2t + 4$. 

\vspace*{-1mm}
In this paper, we work with graphs on which the only possible collisions are meet or head-on collisions. This assumption is a mild one: For example, a 2D grid with unit edge lengths is such a graph for robots with radii of no more than $\sqrt{2}/4$. As a last note, our formulation allows multiple robots to move at the same time step as long as no collision occurs. On a graph, this allows robots on any cycle to ``rotate''.

\vspace*{-1mm}
\subsection{A Motivating Example}
\begin{figure}[htp]
\begin{center}
  \begin{tabular}{ccc}
    \includegraphics[width=0.08\textwidth]{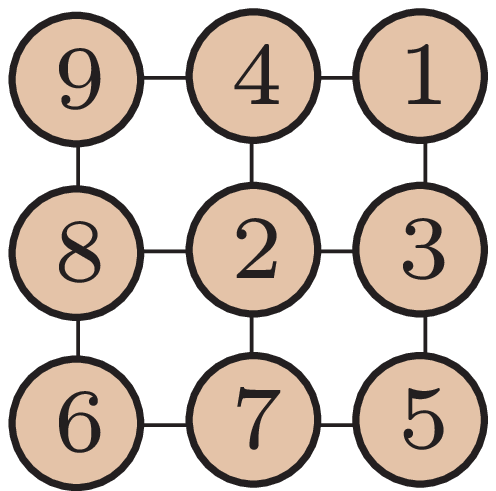} & \hspace{10mm} &
    \includegraphics[width=0.08\textwidth]{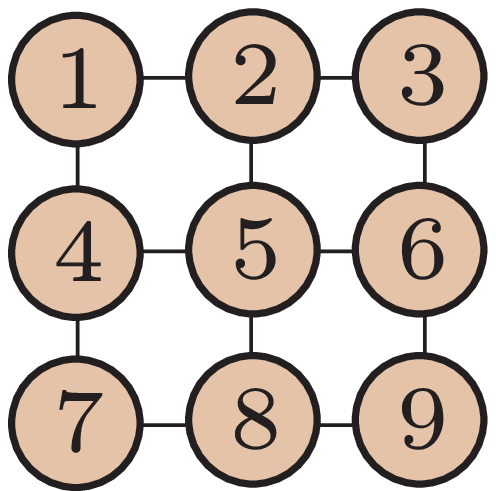}  \\
    (a) && (b)\\
  \end{tabular}
\end{center}
\vspace*{-1mm}
\caption{\label{fig:example} a) A 9-puzzle problem. b) The desired goal state.}
\end{figure}
\begin{figure}[htp]
\begin{center}
    \includegraphics[width=0.35\textwidth]{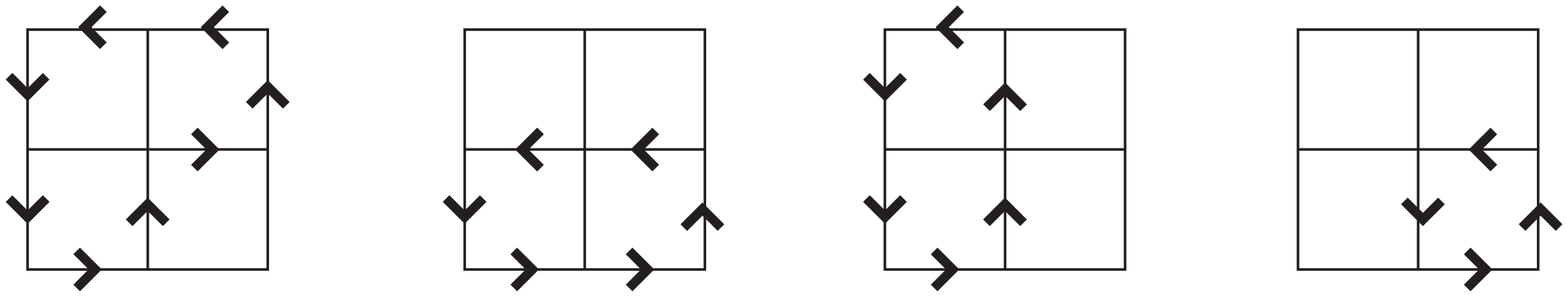} 
\end{center}
\vspace*{-1mm}
\caption{\label{fig:puzzle-8-sol} A 4-step solution from our algorithm. The directed edges show the moving direction of the robots at the tail of the edges.}
\end{figure}

\vspace*{-1mm}
To better characterize what we solve in this paper, look at the example in Fig. \ref{fig:example}. We call this problem a 9-puzzle, which is a variant of the 15-puzzle \cite{RatWar90}; it is also related to the ``H'' example in \cite{LavHut98b}. Given the robots as numbered in Fig. \ref{fig:example}(a), we want to get them into the {\em state} ({\em configuration} is also used in this paper to refer to the same, depending on the context) given in Fig. \ref{fig:example}(b) (such a configuration is often referred to as {\em row major} ordering). Coming up with a feasible solution for such a highly constrained problem is non-trivial, let alone solving it with an optimality guarantee. The time optimal algorithm we present in this paper solves this problem instance under 0.1 second. The solution is given in Fig. \ref{fig:puzzle-8-sol}. The time optimality of the solution is evident: It takes at least four steps for robot 9 to reach its goal. 
\vspace*{-1mm}
\section{Multi-robot Path Planning and Multiflow}\label{sec:planning-and-flow}
\subsection{Network Flow}
In this subsection we provide a summary of the network flow problem formulation pertinent to the introduction of our algorithm. For surveys on network flow, see \cite{Aro89, ForFul62}. A {\em network} $\mathcal N = (G, c_1, c_2, S)$ consists of a directed graph $G = (V, E)$ with $c_1, c_2: E \to \mathbb Z^+$ as the maps defining the capacities and costs on edges, respectively, and $S \subset V$ as the set of sources and sinks. We let $S = S^+ \cup S^-$, with $S^+$ denoting the set of sources and $S^-$ denoting the set of sink vertices. For a vertex $v \in V$, let $\delta^+(v)$ (resp. $\delta^-(v)$) denote the set of edges of $G$ going to (resp. leaving) $v$. A feasible (static) $S^+, S^-$-flow on this network $\mathcal N$ is a map $f: E \to \mathbb Z^+$ that satisfies edge capacity constraints,
\begin{equation}\label{c1}
\forall e \in E, \quad f(e) \le c_1(e),
\end{equation}
the flow conservation constraints at non terminal vertices,
\begin{equation}\label{c2}
\forall v \in V \backslash S, \quad \displaystyle\sum_{e\in \delta^+(v)} f(e)\,\,\, - \sum_{e\in \delta^-(v)} f(e) = 0,
\end{equation}
and the flow conservation constraints at terminal vertices,
\begin{equation}\label{flow-value}
\begin{array}{ll}
F(f) &= \displaystyle\sum_{v \in S^+} (\sum_{e\in \delta^-(v)} f(e)\,\,\, - \sum_{e\in \delta^+(v)} f(e)) \\ 
& = \displaystyle\sum_{v \in S^-} (\sum_{e\in \delta^+(v)} f(e)\,\,\, - \sum_{e\in \delta^-(v)}f(e)).
\end{array}
\end{equation}
The quantity $F(f)$ is called the {\em value} of the flow $f$. The classic (single-commodity) {\em maximum flow} problem asks the question: Given a network $\mathcal N$, what is the maximum $F(f)$ that can be pushed through the network? The {\em minimum cost maximum flow} problem further requires the flow to have minimum total cost among all maximum flows. That is, we want to find a flow among all maximum flows that also minimizes the quantity
\begin{equation}\label{min-cost-max-flow}
\sum_{e \in E} c_2(e)\cdot f(e).
\end{equation}

The above formulation concerns a single commodity, which corresponds to all robots being inter exchangeable. For $\mpp$, the robots are not inter exchangeable and must be treated as different commodities. {\em Multi-commodity flow} or {\em multiflow} captures the problem of flowing different types of commodities through a network. Instead of having a single flow function $f$, we have a flow function $f_i$ for each commodity $i$. The constraints (\ref{c1}), (\ref{c2}), and (\ref{flow-value}) become
\begin{equation}\label{c1m}
\forall i, \forall e \in E, \quad \sum_i \,\, f_i(e) \le c_1(e),
\end{equation}
\begin{equation}\label{c2m}
\forall \, i, \forall \, v \in V \backslash S, \quad \displaystyle\sum_{e\in \delta^+(v)} f_i(e)\,\,\, - \sum_{e\in \delta^-(v)} f_i(e) = 0,
\end{equation}
\begin{equation}\label{flow-value-m}
\begin{array}{lll}
\forall i, & &\displaystyle\sum_{v \in S^+} (\sum_{e\in \delta^-(v)} f_i(e)\,\,\, - \sum_{e\in \delta^+(v)} f_i(e)) \\ 
&=& \displaystyle\sum_{v \in S^-} (\sum_{e\in \delta^+(v)} f_i(e)\,\,\, - \sum_{e\in \delta^-(v)}f_i(e)).
\end{array}
\end{equation}
Again, maximum flow and minimum cost flow problems can be posed for a multiflow setup.

\subsection{Equivalence between $\mpp$ and multiflow}
\begin{figure}[htp]
\begin{center}
  \begin{tabular}{cc}
    \includegraphics[height=0.16\textwidth]{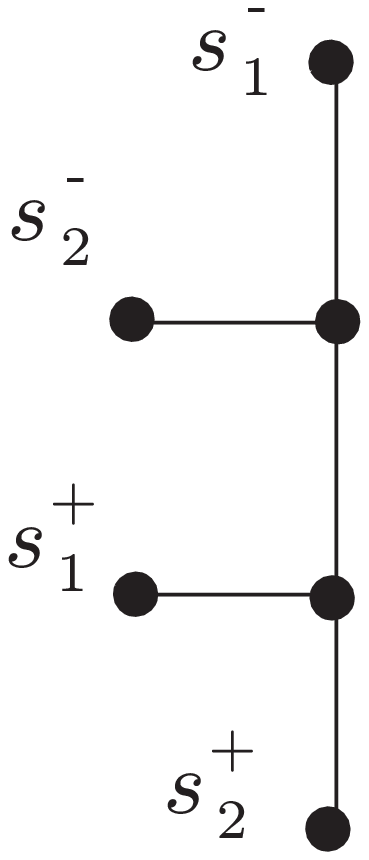} & 
    \includegraphics[height=0.16\textwidth]{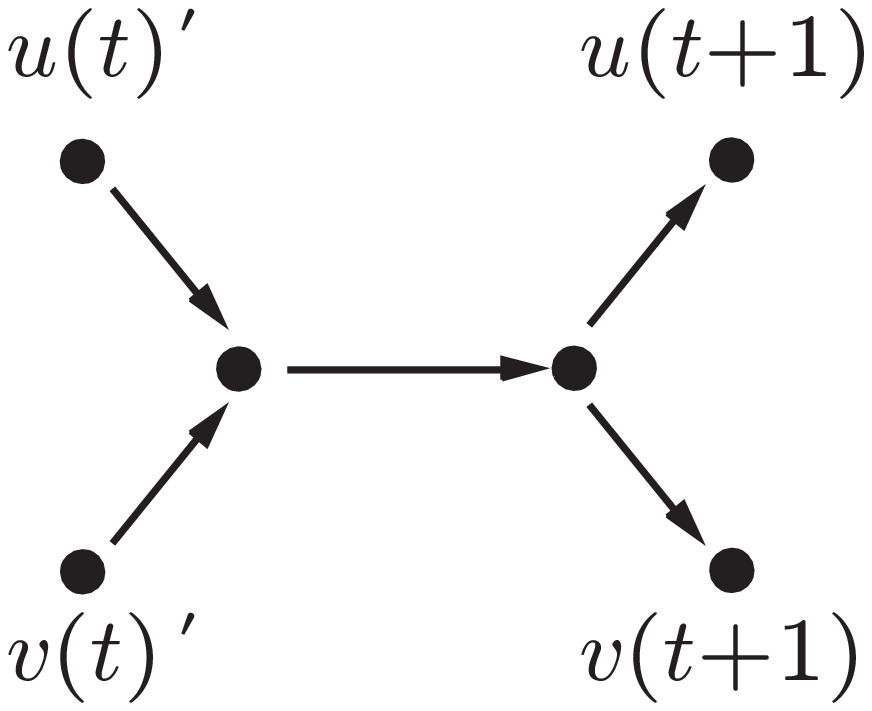}  \\
    (a) & (b)\\
  \end{tabular}
\end{center}
\vspace*{-2mm}
\caption{\label{fig:pimpp} a) A simple $G$. b) A gadget for splitting an undirected edge through time steps.}
\end{figure}

\begin{figure}[htp]
\begin{center}
    \includegraphics[width=0.35\textwidth]{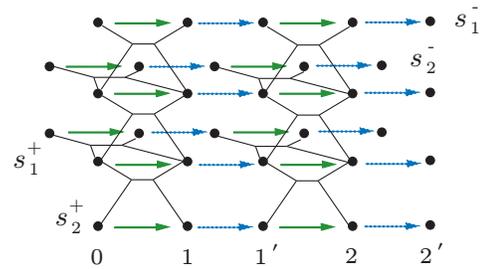}
\end{center}
\vspace*{-4mm}
\caption{\label{fig:pimpp-n} The time-expanded network ($T = 2$).}
\end{figure}

Viewing robots as commodities, we may connect $\mpp$ and multiflow. This relationship (Theorem \ref{t:mpp}) was stated in \cite{YuLav12WAFR-A} without full proof, which is provided here for completeness. To make the presentation clear, we use as an example the simple graph $G$ in Fig. \ref{fig:pimpp}(a), with initial locations $\{s_i^+\}, i = 1, 2$ and goal locations $\{s_i^-\}, i = 1, 2$. An instance of Problem \ref{pimpp} is given by $(G, \{r_1, r_2\}, x_I: r_i \mapsto s^+_i, x_G: r_i \mapsto s^-_i)$. We now convert this problem to a network flow problem, $\mathcal N' = (G', c_1, c_2, S^+ \cup S^-)$. Given the graph $G$ and a natural number 
\vspace*{-2mm}
$T$, we create $2T+1$ copies of vertices from $G$, with indices $0, 1, 1', \ldots$, as shown in Fig. \ref{fig:pimpp-n}. For each vertex $v \in G$, denote these copies $v(0) = v(0)', v(1), v(1)', v(2), \ldots, v(T)'$. For each edge $(u, v) \in G$ and time steps $t, t+1$, $0 \le t < T$, add the gadget shown in Fig. \ref{fig:pimpp}(b) between $u(t)', v(t)'$ and $u(t+1), v(t+1)$ (arrows from the gadget are omitted from Fig. \ref{fig:pimpp-n} since they are small). For the gadget, we assign unit capacity to all edges, unit cost to the horizontal middle edge, and zero cost to the other four edges. This gadget ensures that two robots cannot travel in opposite directions on an edge in the same time step. To finish the construction of Fig. \ref{fig:pimpp-n}, for each vertex $v \in G$, we add one edge between every two successive copies (i.e., we add the edges $(v(0),v(1)), (v(1), v(1)'), \ldots, (v(T), v(T)')$). These correspond to the green and blue edges in Fig. \ref{fig:pimpp-n}. For all green edges, we assign them unit capacity and cost; for all blue edges, we assign them unit capacity and zero cost. 
\vspace*{-1mm}
\vspace*{-1mm}

Fig. \ref{fig:pimpp-n} (with the exception of edges $e_1$ and $e_2$, which are not relevant until Section \ref{sec:algorithm}), called a {\em time-expanded network} \cite{Aro89}, is the desired $G'$. For the set $S$, we may simply let $S^+ = \{v(0): v \in \{s^+_i\} \}$ and $S^- = \{v(T)': v \in \{s^-_i\}\}$. The network $\mathcal N' = (G', c_1, c_2, S^+ \cup S^-)$ is now complete; we have reduced Problem \ref{pimpp} to an integer maximum multiflow problem on $\mathcal N'$ with each robot from $R$ as a single type of commodity. 

\begin{theorem}\label{t:mpp}Given an instance of Problem \ref{pimpp} with input parameters $(G, R, x_I, x_G)$, there is a bijection between its solutions (with maximum number of time steps up to $T$) and the integer maximum multiflow solutions of flow value $n$ on the time-expanded network $\mathcal N'$ constructed from $(G, R, x_I, x_G)$ with $T$ time steps. 
\end{theorem}
{\sc Proof.} (Injectivity) Assume that $P = \{p_1, \ldots, p_n\}$ is a solution to an instance of Problem \ref{pimpp}. For each $p_i$ and every time step $t = 0, \ldots, T$, we mark the copy of $p_i(t)$ and $p_i(t)'$ (recall that $p_i(t)$ corresponds to a vertex of $G$) at time step $t$ in the time-expanded graph $G'$. Connecting these vertices of $G'$ sequentially (there is only one way to do this) yields one unit of flow $f_i$  on $\mathcal N'$ (after connecting to appropriate source and sink vertices in $S^+, S^-$, which is trivial). It is straightforward to see that if two paths $p_i, p_{j}$ are not in collision, then the corresponding flows $f_i, f_j$ on $\mathcal N'$ are vertex disjoint paths and therefore do not violate any flow constraint. Since any two paths in $P$ are not in collision, the corresponding set of flows $\{f_1, \ldots, f_n\}$ is feasible and maximal on $\mathcal N'$. 

(Surjectivity) Assume that $\{f_1,\ldots, f_n\}$ is a integer maximum multiflow on the network $\mathcal N'$ with $|f_i| =1$. First we establish that any pair of flows $f_i, f_j$ are vertex disjoint. To see this, we note that $f_i, f_j$ (both are unit flows) cannot share the same source or sink vertices due to the unit capacity structure of $\mathcal N'$ enforced by the blue edges. If $f_i, f_j$ share some non-sink vertex $v$ at time step $t > 0$, both flows then must pass through the same blue edge (see Fig. \ref{fig:pimpp}(b)) with $v$ being either the head or tail vertex, which is not possible. Thus, $f_i, f_j$ are vertex disjoint on $\mathcal N'$. We can readily convert each flow $f_i$ to a corresponding path $p_i$ (after deleting extra source vertex, sink vertices, vertices in the middle of the gadgets, and tail vertices of blue edges) with the guarantee that no $p_i, p_j$ will collide due to a meet collision. By construction of $\mathcal N'$, the gadget we used ensures that a head-on collision is also impossible. The set $\{p_1, \ldots, p_n \}$ is then a solution to Problem \ref{pimpp}. ~\qed

\subsection{Accommodating other formulations}
\begin{figure}[htp]
\begin{center}
\includegraphics[height=0.16\textwidth]{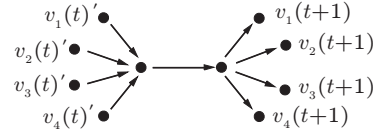}  
\end{center}
\vspace*{-4mm}
\caption{\label{fig:gadget2} A gadget for allowing diagonal crossings.}
\end{figure}

Our network flow based approach for encoding the $\mpp$ problem is fairly general; we illustrate this using two examples. The first is the grid world formulation from \cite{StaKor11}, which allows (single) diagonal crossings. That is, for vertices $v_1, \ldots, v_4$ on the four corners of a square cell with $v_1, v_3$ and $v_2, v_4$ diagonal to each other, respectively, it is possible for a robot to move from $v_1$ to $v_3$ provided that $v_3$ is unoccupied and the $v_2$-$v_4$ diagonal is not used in the same time step. To include this constraint in the ILP model, we may simply add the gadget structure in Fig. \ref{fig:gadget2} to the time-expanded network construction. The inclusion of the gadget will allow a single diagonal crossing; the extra paths do not create an issue since no two robots can go through a single vertex at the same time step (enforced by the blue dotted edges in Fig. \ref{fig:pimpp-n}). 
\vspace*{-2mm}
\vspace*{-2mm}

For a second example, in some $\mpp$ formulations, head-on collisions may be allowed. For instance, two adjacent CPUs may exchange two units of data in parallel but no single CPU may hold multiple units of data. To allow this, we simply do not use the gadget from Fig. \ref{fig:pimpp}(b) when the time-expanded network is constructed.  

\section{Algorithmic Solutions for Optimal Multi-robot Path Planning}\label{sec:algorithm}

Given the time-expanded network $\mathcal N' = (G', c_1, c_2, S^+ \cup S^-)$, it is straightforward to create an integer linear programming (ILP) model with different optimality objectives. We investigate two objectives in this section: Time optimality or makespan (the time when the last robot reaches its goal) and distance optimality (the total distance traveled by all robots). 

\subsection{Time optimality}
Time optimal solutions to Problem \ref{pimpp} can be obtained using a maximum multiflow formulation. As a first step, we introduce a set of $n$ {\em loopback} edges to $G'$ by connecting each pair of corresponding goal and start vertices in $S$, from the goal to the start. For convenience, denote these loopback edges as $\{e_1, \ldots, e_n\}$ (e.g., edges $e_1, e_2$ in Fig. \ref{fig:pimpp-n}). These edges have unit capacity and zero cost. Next. for each edge $e_j \in G'$, create $n$ binary variables $x_{1, j}, \ldots, x_{n,j}$ corresponding to the flow through that edge, one for each robot. $x_{i, j} = 1$ if and only if robot $r_i$ passes through $e_j$ in $G'$. The variables $x_{i,j}$'s must satisfy two edge capacity constraints and one flow conservation constraint, 
\begin{equation}\label{to1}
\begin{array}{cc}
\forall\, e_j \in G', & \displaystyle\sum_{i=1}^n x_{i,j} \le 1\\
\forall\, 1 \le i, j \le n, \,i \ne j, & \displaystyle x_{i, j} = 0, 
\end{array}
\end{equation}
\begin{equation}\label{to2}
\forall\, v \in G' \textrm{ and } 1 \le i \le n, \displaystyle\sum_{e_j \in \delta^+(v)} x_{i,j} = \sum_{e_j \in \delta^-(v)} x_{i,j}.
\end{equation}
The objective function is 
\begin{equation}\label{to3}
\max \sum_{1 \le i \le n} x_{i,i}.
\end{equation}

For each fixed $T$, the solution to the above ILP problem equaling $n$ means that a feasible solution to Problem \ref{pimpp} is found. We are to find the minimal $T$ that yields such a feasible solution. To do this, we start with $T$ being the maximum over all robots the shortest possible path length for each robot, ignoring all other robots. We then build the ILP model for this $T$ and test for a feasible solution. If the model is not feasible, we increase $T$ and try again. The first feasible $T$ is the optimal $T$. The robots' paths can be extracted based on the proof of Theorem \ref{t:mpp}. The algorithm is complete: Since the problem is discrete, there is only a finite number of possible states. Therefore, for some sufficiently large $T$, there must either be a feasible solution or we can pronounce that none can exist. Calling this algorithm \tompp\, (time optimal $\mpp$), we have shown the following.
\begin{proposition}\label{p:time}Algorithm \tompp\, is complete and returns a solution with minimum makespan to Problem \ref{pimpp} if one exists.
\end{proposition}

\subsection{Distance optimality}

Distance optimality objective can be encoded using minimum cost maximum multiflow. Constraints (\ref{to1}) and (\ref{to2}) remain; to force a maximum flow, let $x_{i,i} = 1$ for $1 \le i \le n$. The objective is given by
\begin{equation}\label{to4}
\min \sum_{e_j \in G', j > n,\, 1 \le i \le n} c_2(e_j) \cdot x_{i,j}.
\end{equation}

The value given by (\ref{to4}), when feasible, is the total distance of all robots' paths. Let $T_t$ denote the optimal $T$ produced by \tompp\,(if one exists), then a distance optimal solution exists in a time-expanded network with $T = nT_t$ steps. Calling this algorithm \dompp\, (distance optimal $\mpp$), we have 
\begin{proposition}Algorithm \dompp\, is complete and returns a solution with minimum total path length to Problem \ref{pimpp} if one exists.
\end{proposition}

Due to the large number of steps needed in the time-expanded network, \dompp, in its current form, is not very fast in solving problems with many robots. Therefore, our evaluation in this paper focuses on \tompp\, which, on the other hand, is fairly fast in solving some very difficult problems. \dompp, however, still proves useful in providing time optimal and near distance optimal solutions using the outputs of \tompp, as shown in Subsection \ref{subsec:dompp}. 


\section{Properties of the $n^2$-puzzle}\label{sec:puzzle}
The example problem from Fig. \ref{fig:example} easily extends to an $n \times n$ grid; we call this class of problems the $n^2$-puzzle. Such problems are highly coupled: No robot can move without at least three other robots moving at the same time. At each step, all robots that move must move synchronously in the same direction (per cycle) on one or more disjoint cycles (see e.g., Fig. \ref{fig:puzzle-8-sol}). To put into perspective the computational results on $n^2$-puzzles that follow, we make a characterization of the state structure of the $n^2$-puzzle for $n \ge 3$ (the case of $n=2$ is trivial). 
\begin{figure}[htp]
\begin{center}
    \includegraphics[width=0.35\textwidth]{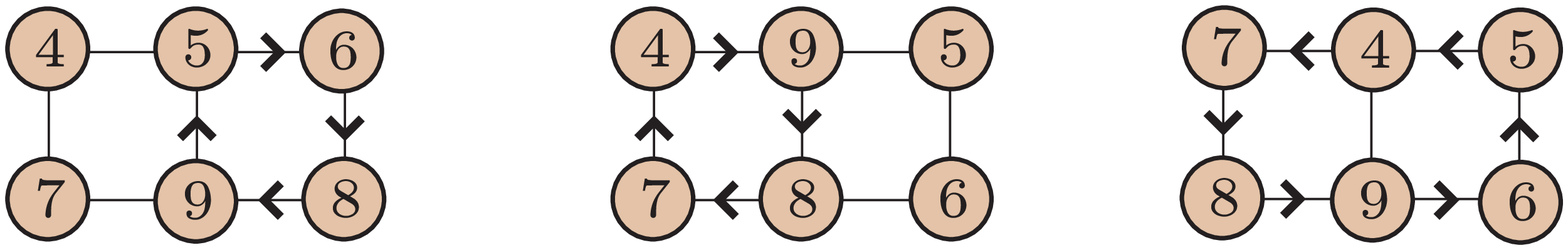} 
\end{center}
\vspace*{-2mm}
\caption{\label{fig:6-puzzle} A 3-step procedure for exchanging robots 8 and 9.}
\end{figure}
\vspace*{-2mm}

\begin{proposition}\label{p:state}All states of the 9-puzzle are connected via legal moves. 
\end{proposition}
{\sc Proof}. We show that any state of a 9-puzzle can be moved into the state shown in Fig. \ref{fig:example}(b). From any state, robot 5 can be easily moved into the center of the grid. We are left to show that we can exchange two robots on the border without affecting other robots. This is possible due to the procedure illustrated in Fig. \ref{fig:6-puzzle}. 
~\qed

\begin{figure}[htp]
\begin{center}
    \includegraphics[width=0.35\textwidth]{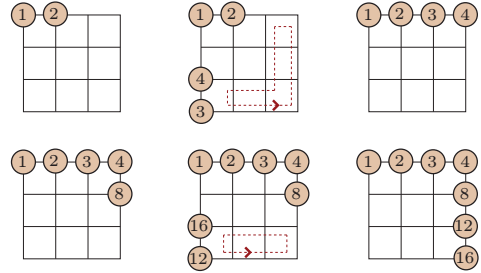} 
\end{center}
\vspace*{-2mm}
\caption{\label{fig:16-puzzle} A solution scheme for solving top/left sides of the 16-puzzle.}
\end{figure}

Larger puzzles can be solved recursively: We may first solve the top and right side of the puzzle and then the left over smaller square puzzle. For a 16-puzzle, Fig. \ref{fig:16-puzzle} outlines the procedure, consisting of six main steps:
\begin{enumerate}
\item Move robots 1 and 2 to their respective goal locations, one robot at a time (first 1, then 2). 
\item Move robots 3 and 4 (first 3, then 4) to the lower left corner (top-middle figure in Fig. \ref{fig:16-puzzle}). 
\item Move robots 3 and 4 to their goal location together via counterclockwise rotation along the cycle indicated in the top-middle figure in Fig. \ref{fig:16-puzzle}. 
\item Move robot 8 to its goal location. 
\item Move robots 12 and then 16 to the lower left corner. 
\item Rotate robots 12 and 16 to their goal locations. 
\end{enumerate}
\vspace*{-2mm}

It is straightforward to see that larger puzzles can be solved similarly. We have thus outlined the essential steps for proving Proposition \ref{c:state} below; a more generic proof can be written using generators of permutation groups, which we omit here due to its length. Proposition \ref{c:state} implies that, for $n \ge 3$, all instances of $n^2$-puzzles are solvable. The constructive proofs of Proposition \ref{p:state} and \ref{c:state} lead to recursive algorithms for solving any $n^2$-puzzle (clearly, the solution is not time/distance optimal in general). 

\begin{proposition}\label{c:state}All states of an $n^2$-puzzle, $n \ge 3$ are connected via legal moves. 
\end{proposition}
\begin{corollary}\label{c:solvable}All instances of the $n^2$-puzzle, $n \ge 3$, are solvable. 
\end{corollary}

By Proposition \ref{c:state}, since all states of a $n^2$-puzzle for $n \ge 3$ are connected via legal moves, the state space of searching an $n^2$-puzzle equals $n^2$ {\em factorial}. For  16-puzzle and 25-puzzle, $16! > 10^{13}, 25! > 10^{25}$. Large state space is one of the three reasons that make finding a time optimal solution to the $n^2$-puzzle a difficult problem. The second difficulty comes from the large branching factor at each step. For a 9-puzzle, there are 13 unique cycles, yielding a branching factor of 26 (clockwise and counterclockwise rotations). For the 16-puzzle, the branching factor is around 500. This number balloons to over $10^4$ for the 25-puzzle. This suggests that on typical commodity personal computer hardware (assuming a 1GHz processor), a baisc breadth first search algorithm will not be able to go beyond depth of 3 for the 16-puzzle and depth 2 for the 25-puzzle in reasonable amount of time. Moreover, enumerating these cycles is a non-trivial task. The third difficulty is the lack of obvious heuristics: Manhattan distances of robots to their respective goals prove to be a bad one. For example, given the initial configuration as that in Fig. \ref{fig:example}(a), the first step in the optimal plan from Fig. \ref{fig:puzzle-8-sol} gets robots 1, 3, 4, 6, 8, 9 closer to their respective goals while moving robots 2, 7 farther. On the other hand, rotating counterclockwise along the outer cycle takes robots 1, 3, 4, 5, 6, 8, 9 closer and only moves robot 7 farther. However, if we instead take this latter first step, the optimal plan afterwards will take 5 more steps. 

\section{Solutions and Evaluation}\label{sec:evaluation}

Our experimentation in this paper focuses on \tompp\, with the main goal being evaluating the comparative efficiency of our approach rather than pushing for best computational performance. As such, our implementation is Java based and did not directly take advantage of multi-core technology. We note that, Gurobi, the ILP solver used in our implementation, can engage multiple cores automatically for hard problems. We ran our code on an Intel Q6600 quad-core machine with a 4GB JavaVM. 

\subsection{Time optimal solution to $n^2$-puzzles}
The first experiment we performed was evaluating the efficiency of the algorithm \tompp\, for finding time optimal solutions to the $n^2$-puzzle for $n = 3, 4, 5,$ and $6$. We ran Algorithm \tompp\, on 100 randomly generated $n^2$-puzzle instances for $n = 3, 4, 5$. For the 9-puzzle, computation on all instances completed successfully with an average computation time of 1.36 seconds per instance. To compare the computational result, we implemented a (optimal) BFS algorithm. The BFS algorithm is heavily optimized: For example, cycles of the grid are precomputed and hard coded to save computation time. Since the state space of the 9-puzzle is small, the BFS algorithm is capable of optimally solving the same set of 9-puzzle instances with an average computation time of about 0.89 seconds per instance. 

Once we move to the 16-puzzle, the power of general ILP solvers becomes evident. \tompp\, solved all 100 randomly generated 16-puzzle instances with an average computation time of 18.9 seconds. On the other hand, the BFS algorithm with a priority queue that worked for the 9-puzzle ran out of memory after a few minutes. As our result shows that an optimal solution for the 16-puzzle generally requires 6 time steps, it seems natural to also try bidirectional search, which cuts down the total number states stored in memory. To complete such a search, one side of the bidirectional search generally must reach a depth of 3, which requires storing about $3 \times 10^7$ states, each taking 64 bits of memory. This turns out to be too much for a 4GB JavaVM: A bidirectional search ran out of memory after about 10 minutes in general. To be sure, we also coded part of the same search algorithm in C++ with STL. Reaching a search depth 3 on one side takes about a minute with a memory footprint of 1.5GB, suggesting a minimum running time of more than one minute. 

\begin{figure}[htp]
\begin{center}
    \includegraphics[width=0.12\textwidth]{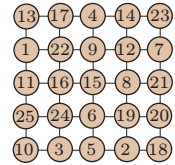} 
\end{center}
\vspace*{-2mm}
\caption{\label{fig:25-puzzle} An instance of a 25-puzzle problem solved by \tompp.}
\end{figure}
\vspace*{-2mm}

\begin{figure}[htp]
\begin{center}
    \includegraphics[width=0.35\textwidth]{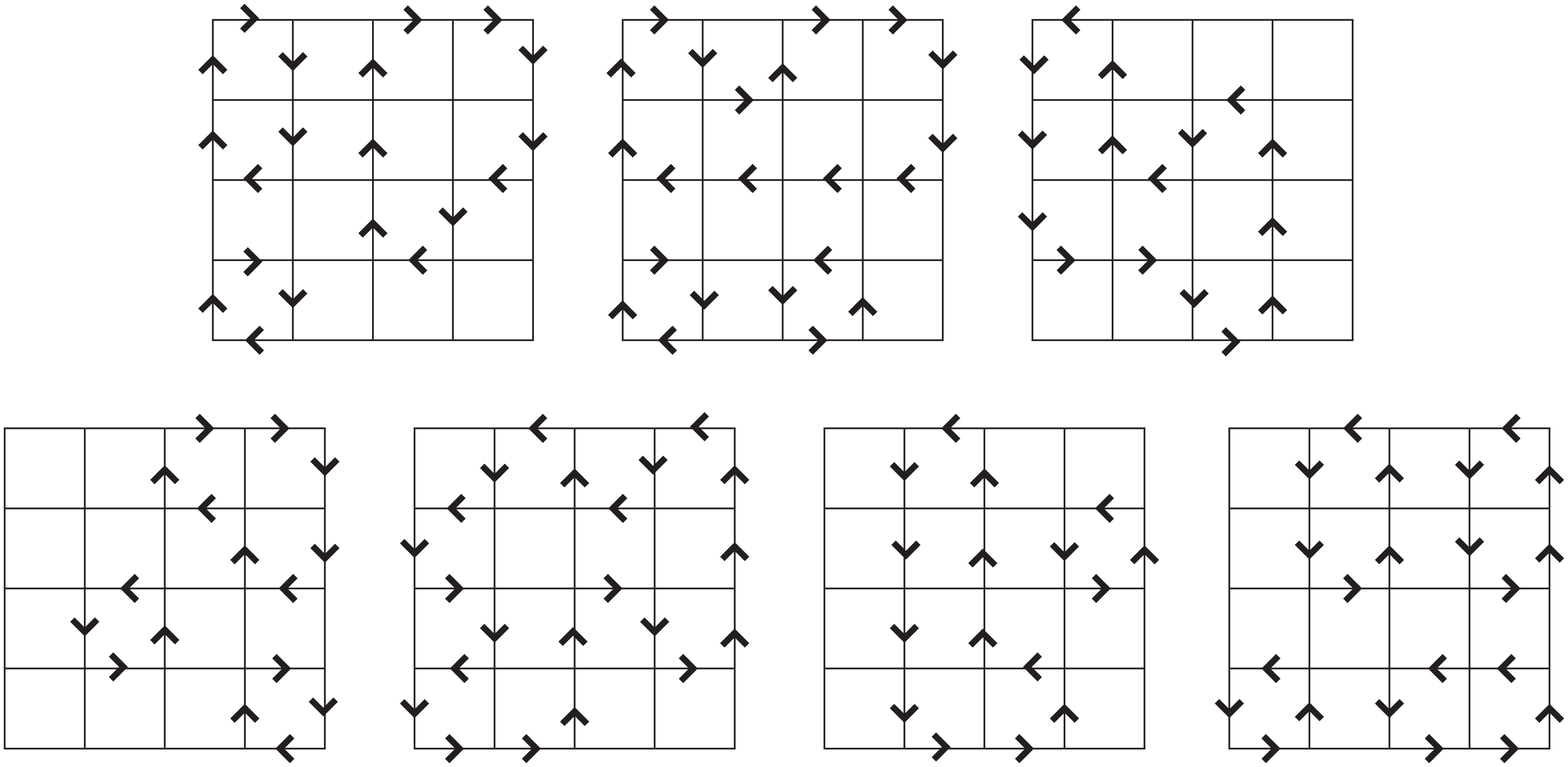} 
\end{center}
\caption{\label{fig:25-puzzle-sol} An optimal 7-step solution (from left to right, then top to bottom) to the 25-puzzle problem from Fig. \ref{fig:25-puzzle}, by \tompp\, in about 30 minutes.}
\end{figure}

For the 25-puzzle, without a good heuristic, bidirectional search cannot explore a tiny fraction of the fully connected state space with about $10^{25}$ states. On the other hand, \tompp\, again consistently solves the 25-puzzle, with an average computational time under 2 hours over 100 randomly created problems. Fig. \ref{fig:25-puzzle} shows one of the solved instances with a 7-step solution given in Fig. \ref{fig:25-puzzle-sol}. Note that 7 steps is obviously the least 
possible since it takes at least 7 steps to move robot 10 to its desired goal. We also briefly tested \tompp\, on the 36-puzzle. While we had some success here, \tompp\, generally does not seem to solve a randomly generated instance of the 36-puzzle within 24 hours, which has $3.7 \times 10^{41}$ states and a branching factor of well over $10^6$. 
\subsection{Time optimal solutions for grid graphs}
\begin{figure}[htp]
\begin{center}
    \includegraphics[width=0.4\textwidth]{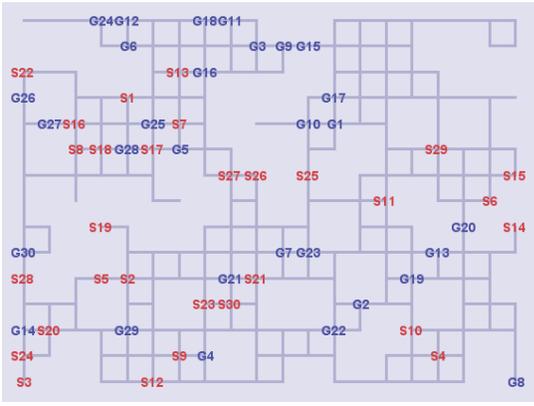} 
\end{center}
\vspace*{-2mm}
\caption{\label{fig:20x15} A $20 \times 15$ grid with 20\% verices removed (modeling obstacles) and 30 start/goal pairs. The start locations are marked with strings beginning with ``S'' and the goal locations are marked with strings beginning with ``G''.}
\end{figure}

For problems in which not all graph vertices are occupied by robots, \tompp\, can handle much larger instances. In a first set of tests on this subject, a grid size of $20 \times 15$ is used with varying percentage of obstacles (simulated by removed vertices) and robots for evaluating the effect of these factors. A typical set up is illustrated in Fig. \ref{fig:20x15}. The computation time (in seconds) and the average number of optimal time steps (in parenthesis) are listed in Table \ref{tab:20x15}. The numbers are 
\begin{table}[htp]
\begin{center}
	 \caption{\label{tab:20x15}}
\vspace*{-1mm}
	 \begin{tabular}{cccccc}
   \hline\hline
	 \multirow{2}*{\% obs} & \multicolumn{5}{c}{Number of robots} \\
	 \cline{2-6}
	  & 10 & 20 & 30 & 40 & 50 \\
	 \hline
	 5 & 2.5(22) & 7.3(24) & 16.7(27) & 23.6(26) & 70.7(27) \\
	 \hline
	 10 & 2.1(21) & 7.8(24) & 13.1(26) & 20.4(26) & 48.6(26) \\
	 \hline
	 15 & 3.9(25) & 6.2(24) & 13.8(26) & 32.8(27) & 126(28) \\
	 \hline
	 20 & 2.4(24) & 7.7(27) & 21.9(28) & 39.3(26) & 173(27) \\
	 \hline
	 25 &  2.7(27) & 8.1(28) & 24.8(30) & 68.0(28) & $253(30)^4$ \\
	 \hline
	 30 &  3.0(31) & $29.9(34)^9$ & $234(44)^5$ & $80.6(29)^3$ & N/A \\
	 \hline\hline
	 \end{tabular}
\vspace*{-2mm}
\end{center}
\end{table}
averages over 10 randomly created instances. For each run, a maximum of 1000 seconds is allowed (such limits, somewhat arbitrary, were chosen to manage the expected running time of the entire set of experiments; our complete algorithms should terminate eventually). Entries with superscript numbers suggest the 10 runs did not all finish within the given time. The superscript numbers represent the successful runs on which the statistics were computed. ``N/A'' means no instance finished within the allowed time. From the results, we observe that the percentage of randomly placed obstacles does not affect the problem difficulty, as measured by computational time, in a monotonic way. On one hand, more obstacles remove more vertices from the grid, making the problem size smaller, reducing the computational difficulty. On the other hand, as more obstacles are introduced, the reduced connectivity of the graph makes the problem harder. In particular, the $20 \times 15$ grid setting suddenly becomes a hard problem with 30\% obstacles. The difficulty is also reflected by the average number of steps in an optimal solution: Longer time means reduced availability of alternative paths. 
\begin{table}[htp]
\begin{center}
	 \caption{\label{tab:32x32p}}
\vspace*{-1mm}
	 \begin{tabular}{cccccc}
   \hline\hline
	 \multirow{2}*{\% obs} & \multicolumn{5}{c}{Number of robots} \\
	 \cline{2-6}
	  & 10 & 20 & 30 & 40 & 50 \\
	 \hline
	 20 & 14.4(41) & 34.6(45) & 43.7(44) & 87.5(47) & $402(49)^9$ \\
	 \hline\hline
	 \end{tabular}
\vspace*{-2mm}
\end{center}
\end{table}

In a second test on even larger problems, $32 \times 32$ grids with 20\% obstacles were tried. For between 10 and 50 robots with an increment of 10, 10 random instances each were created; each instance is allowed to run a maximum of half an hour. The statistics, similarly composed as that in Table \ref{tab:20x15}, is listed in Table \ref{tab:32x32p}. We observe that the problem is similar in difficulty to the $20 \times 15$ grid setting with 25\% obstacles, but much simpler than that with 30\% obstacles. 
\subsection{Distance optimality of time optimal solutions}\label{subsec:dompp}

Although \dompp\, is not yet practical for computing distance optimal solutions alone, it can be used for computing distance optimal solutions for a fixed time expansion length $T$. That is, we first find a time optimal solution, which gives us the smallest time-expanded network containing feasible solutions. We then run \dompp\, on this network. For evaluation, we used the same $20 \times 15$ instances with 5-25\% obstacles and 10-30 robots (\dompp\, could not finish most instances with 30\% obstacles or 40+ robots in 200 seconds, the cutoff time). We used the first 5 of every 10 instances for each obstacle/robot combination. For each fixed number of obstacles, instances of different numbers of robots are combined. The result is listed in Table \ref{tab:20x15d}. We allow \dompp\, to run for at most 200 seconds per instance. Note that unlike \tompp, even when \dompp\, does not find the optimal solution, it generally produces feasible solution which sometimes is a near optimal solution. These are included in the result. ``Time'' entires are average time, in seconds, used by \dompp. ``Disjoint'' entries are the average path lengths for all robots if we were to plan each shortest path ingoring other robots. The distance optimal solutions must produce a length no less than this. The next two lines are average path lengths from \tompp\, and \dompp\, algorithms. As we can see, \tompp\, alone yields path length 50\% than optimal; \dompp, on the other hand, provided time optimal solutions that are near distance optimal ($< 1\%$ difference). For more than half of the instances, \dompp\, produced true distance optimal solutions. In fact, \dompp\, produced true distance optimal solutions for 42 out of the 45 instances with 5-15\% obstacles.
\begin{table}[htp]
\begin{center}
	 \caption{\label{tab:20x15d}}
\vspace*{-1mm}
	 \begin{tabular}{cccccccccc}
   \hline\hline
	 \multirow{2}*{} & \multicolumn{9}{c}{\% obs} \\
	 \cline{2-10}
	  & 5 && 10 && 15 && 20 && 25 \\
	 \hline
	 Time & 26.3 && 23.3 && 42.7 && 57.2 && 81.6 \\
	 \hline
	 Disjoint & 12.20 && 11.75 && 12.03 && 12.80 && 12.84 \\
	 \hline
	 \dompp & 12.20 && 11.75 && 12.05 && 12.85 && 12.92 \\
	 \hline
	 \tompp & 16.47 && 16.60 && 17.59 && 18.83 && 19.33 \\
	 \hline\hline
	 \end{tabular}
\vspace*{-2mm}
\end{center}
\end{table}
\vspace*{-2mm}
\subsection{Using {\sc Tompp} as a generic heuristic}
In the last experiment, we exploit \tompp\, as a {\em generic} heuristic for locally resolving path conflicts for large problem instances. By {\em generic}, we mean that the heuristic is not coded to any specific robot/grid setting. In our algorithm, paths are first planned for single robots (ignoring other robots). Afterwards, the robots are moved along these paths until no further progress can be made. We then detect on the graph where progress are stalled and resolve the conflict locally using \tompp. For every conflict, we apply \tompp\, to its neighborhood of distance 2. The above steps are repeated until a solution is found. The process can be made into a complete algorithm by allowing the local neighborhood to grow gradually. For evaluation, we ran the above algorithm on a $32\times 32$ grid with 20\% obstacles. We allow each instance to run a maximum of 30 seconds. The results, each as an average over 100 runs for a certain number of robots, are listed in Table \ref{tab:32x32} (keep \begin{table}[htp]
\begin{center}
	\caption{\label{tab:32x32}}
	\begin{tabular}{rrrrrrr}
   \hline\hline
	  & \multicolumn{6}{c}{Number of Robots} \\
	 \cline{2-7}
	   & 25 & 50 & 75 & 100 & 125 & 150\\
	 \hline
	 Running time (s) & 0.04 & 0.15 & 0.32 & 1.37 & 3.85 & 10.3 \\
	 \hline
	 Fully solved & 100 & 100 & 100 & 100 & 98 & 95 \\
	 \hline
	 \% goals reached & 100.0 & 100.0 & 100.0 & 100.0 & 99.4 & 98.6 \\
	 \hline\hline
	 \end{tabular}
\vspace*{-2mm}
\end{center}
\end{table}
in mind that our implementation is Java based, which should see a speedup if implemented in C++). While we did not make side-by-side comparisons with the literature due to (seemingly small but) important differences in problem formulation, the computation time and completion rate of our algorithm appear comparable with the state of the art results from other authors. 

\section{Conclusion and Open Problems}\label{sec:conclusion}

In this paper, we introduced a multiflow based ILP algorithm for planning optimal, collision-free paths for multiple robots on graphs. We provided complete ILP algorithms for solving time optimal and distance optimal $\mpp$ problems. Our experiments confirmed that \tompp\, is a feasible method for planning time optimal paths for tightly coupled problems as well as for larger problems with more free space. Moreover, we showed that \tompp\, can serve as a good heuristic for solving large problem instances efficiently. For distance optimality, \dompp, when combined with \tompp, produces time optimal solutions that are often near distance optimal. 

Many interesting open problems on optimal $\mpp$ remain; we mention two here. First, the ILP  algorithms have ample room for performance improvements. On one hand, the ILP model can be make leaner. For example, it is clear that some $x_{i,j}$'s will never be set to 1; these should be removed from the model. On the other hand, our application of the Gurobi solver is fairly rudimentary - we simply feed the model to the solver as a mixed integer program (MIP) without specifying any other optimization options. Therefore, it would not be surprising that tuning the parameters of the solver greatly improves its performance on $\mpp$ problems. Secondly, while \tompp\, could solve hard $\mpp$ problems such as the 25-puzzle, ILP solvers are nevertheless not tailored for such problems. Thus, we expect that tailored methods, such as heuristic based search, to solve problems like $n^2$-puzzles even faster. Looking closely at how ILP solvers work on these problems should provide insights that help building these heuristics. 

\bibliographystyle{plain}
\bibliography{../../../../../references/references,../../../../../references/references2,../../../../../references/publications}

\begin{thebibliography}{10}

\bibitem{Aro89}
J.~E. Aronson.
\newblock A survey on dynamic network flows.
\newblock {\em Annals of Operations Research}, 20(1):1--66, 1989.

\bibitem{BalArk98}
T.~Balch and R.~C. Arkin.
\newblock Behavior-based formation control for multirobot teams.
\newblock {\em IEEE Transaction on Robotics and Automation}, 14(6):926--939,
  1998.

\bibitem{ChoLynHutKanBurKavThr05}
H.~Choset, K.~M. Lynch, S.~Hutchinson, G.~Kantor, W.~Burgard, L.~E. Kavraki,
  and S.~Thrun.
\newblock {\em Principles of Robot Motion: Theory, Algorithms, and
  Implementations}.
\newblock MIT Press, Cambridge, MA, 2005.

\bibitem{ErdLoz86}
M.~A. Erdmann and T.~Lozano-P\'erez.
\newblock On multiple moving objects.
\newblock In {\em Proceedings IEEE International Conference on Robotics \&
  Automation}, pages 1419--1424, 1986.

\bibitem{ForFul62}
L.~R. Ford and D.~R. Fulkerson.
\newblock {\em Flows in Networks}.
\newblock Princeton University Press, New Jersey, 1962.

\bibitem{FoxBurKruThr00}
D.~Fox, W.~Burgard, H.~Kruppa, and S.~Thrun.
\newblock A probabilistic approach to collaborative multi-robot localization.
\newblock {\em Autom. Robots}, 8(3):325--344, June 2000.

\bibitem{HalLatWil00}
D.~Halperin, J.-C. Latombe, and R.~Wilson.
\newblock A general framework for assembly planning: The motion space approach.
\newblock {\em Algorithmica}, 26(3-4):577--601, 2000.

\bibitem{HarNilRap68}
P.~Hart, N.~J. Nilsson, and B.~Raphael.
\newblock A formal basis for the heuristic determination of minimum cost paths.
\newblock {\em IEEE Transactions on Systems Science and Cybernetics},
  4:100--107, 1968.

\bibitem{JenWheEva97}
J.~S. Jennings, G.~Whelan, and W.~F. Evans.
\newblock Cooperative search and rescue with a team of mobile robots.
\newblock In {\em Proceedings IEEE International Conference on Robotics \&
  Automation}, 1997.

\bibitem{Lat91}
J.-C. Latombe.
\newblock {\em Robot Motion Planning}.
\newblock Kluwer, Boston, MA, 1991.

\bibitem{Lav06}
S.~M. LaValle.
\newblock {\em Planning Algorithms}.
\newblock Cambridge University Press, Cambridge, U.K., 2006.
\newblock Also available at http://planning.cs.uiuc.edu/.

\bibitem{LavHut98b}
S.~M. LaValle and S.~A. Hutchinson.
\newblock Optimal motion planning for multiple robots having independent goals.
\newblock {\em IEEE Trans. on Robotics and Automation}, 14(6):912--925,
  December 1998.

\bibitem{LunBer11}
R.~Luna and K.~E. Bekris.
\newblock Push and swap: Fast cooperative path-finding with completeness
  guarantees.
\newblock In {\em Twenty-Second International Joint Conference on Artificial
  Intelligence}, pages 294--300, 2011.

\bibitem{MatNilSim95}
M.~J. Matari\'c, M.~Nilsson, and K.~T. Simsarian.
\newblock Cooperative multi-robot box pushing.
\newblock In {\em Proceedings IEEE/RSJ International Conference on Intelligent
  Robots and Systems}, pages 556--561, 1995.

\bibitem{Nna92}
B.~Nnaji.
\newblock {\em Theory of Automatic Robot Assembly and Programming}.
\newblock Chapman \& Hall, 1992.

\bibitem{PodSuk04}
S.~Poduri and G.~S. Sukhatme.
\newblock Constrained coverage for mobile sensor networks.
\newblock In {\em Proceedings IEEE International Conference on Robotics \&
  Automation}, 2004.

\bibitem{RatWar90}
D.~Ratner and M.~Warmuth.
\newblock The $(n^2-1)$-puzzle and related relocation problems.
\newblock {\em Journal of Symbolic Computation}, 10:111--137, 1990.

\bibitem{RodAma10}
S.~Rodriguez and N.~M. Amato.
\newblock Behavior-based evacuation planning.
\newblock In {\em Proceedings IEEE International Conference on Robotics and
  Automation}, pages 350--355, 2010.

\bibitem{RusDonJen95}
D.~Rus, B.~Donald, and J.~Jennings.
\newblock Moving furniture with teams of autonomous robots.
\newblock In {\em Proceedings IEEE/RSJ International Conference on Intelligent
  Robots and Systems}, pages 235--242, 1995.

\bibitem{Rya08}
M.~R.~K. Ryan.
\newblock Exploiting subgraph structure in multi-robot path planning.
\newblock {\em Journal of Artificial Intelligence Research}, 31:497--542, 2008.

\bibitem{ShuMurBen07}
B.~Shucker, T.~Murphey, and J.~K. Bennett.
\newblock Switching rules for decentralized control with simple control laws.
\newblock In {\em American Control Conference}, July 2007.

\bibitem{Sil05}
D.~Silver.
\newblock Cooperative pathfinding.
\newblock In {\em The 1st Conference on Artificial Intelligence and Interactive
  Digital Entertainment}, pages 23--28, 2005.

\bibitem{SmiEgeHow08}
B.~Smith, M.~Egerstedt, and A.~Howard.
\newblock Automatic generation of persistent formations for multi-agent
  networks under range constraints.
\newblock {\em ACM/Springer Mobile Networks and Applications Journal},
  14(3):322--335, June 2009.

\bibitem{StaKor11}
T.~Standley and R.~Korf.
\newblock Complete algorithms for cooperative pathfinding problems.
\newblock In {\em Twenty-Second International Joint Conference on Artificial
  Intelligence}, pages 668--673, 2011.

\bibitem{Sur09}
P.~Surynek.
\newblock A novel approach to path planning for multiple robots in bi-connected
  graphs.
\newblock In {\em Proceedings IEEE International Conference on Robotics and
  Automation}, pages 3613--3619, 2009.

\bibitem{Sur10}
P.~Surynek.
\newblock An optimization variant of multi-robot path planning is intractable.
\newblock In {\em The Twenty-Fourth AAAI Conference on Artificial Intelligence
  (AAAI-10)}, pages 1261--1263, 2010.

\bibitem{TanPapKum04}
H.~Tanner, G.~Pappas, and V.~Kumar.
\newblock Leader-to-formation stability.
\newblock {\em IEEE Transactions on Robotics and Automation}, 20(3):443--455,
  Jun 2004.

\bibitem{BerOve05}
J.~van~den Berg and M.~Overmars.
\newblock Prioritized motion planning for multiple robots.
\newblock In {\em Proceedings IEEE/RSJ International Conference on Intelligent
  Robots and Systems}, 2005.

\bibitem{BerSnoLinMan09}
J.~van~den Berg, J.~Snoeyink, M.~Lin, and D.~Manocha.
\newblock Centralized path planning for multiple robots: Optimal decoupling
  into sequential plans.
\newblock In {\em Proceedings Robotics: Science and Systems}, 2009.

\bibitem{YuArxiv-1205-5263}
J.~Yu.
\newblock Diameters of permutation groups on graphs and linear time feasibility
  test of pebble motion problems.
\newblock {\em arXiv:1205.5263}, 2012.

\bibitem{YuLav12WAFR-A}
J.~Yu and S.~M. LaValle.
\newblock Multi-agent path planning and network flow.
\newblock In {\em The Tenth International Workshop on Algorithmic Foundations
  of Robotics}, 2012.

\bibitem{Zel92}
A.~Zelinsky.
\newblock A mobile robot exploration algorithm.
\newblock {\em IEEE Transactions on Robotics and Automation}, 8(6):707--717,
  1992.

\end{thebibliography}
\end{document}